\documentclass[conference]{IEEEtran}
\IEEEoverridecommandlockouts
\usepackage{hyperref}
\usepackage{url}
\usepackage{bm}
\usepackage{amsmath,amsfonts}
\usepackage{algorithmicx}
\usepackage{algpseudocode}
\usepackage{algorithm}
\usepackage{array}
\usepackage{xcolor}
\usepackage{subcaption}
\usepackage{stfloats}
\usepackage{url}
\usepackage{verbatim}
\usepackage{graphicx}

\usepackage{enumitem}

\usepackage{cleveref}
\usepackage{bbm}
\usepackage[acronym]{glossaries}
\glsdisablehyper
\usepackage{siunitx}
\usepackage{dsfont}
\usepackage{textgreek}
\usepackage{multirow}
\usepackage{pifont}
\usepackage{calc}
\usepackage{threeparttable}
\usepackage{tabularx}
\usepackage[pscoord]{eso-pic}
\graphicspath{fig}

\DeclareSIUnit\op{Op}
\newlist{AlgoDx}{description}{2}
\setlist[AlgoDx,1]{%
  noitemsep,
  labelindent=\algorithmicindent*\real{1.3},
  font=\normalfont,
  labelsep=\algorithmicindent,
  leftmargin=\algorithmicindent*\real{2.0}
}
\setlist[AlgoDx]{%
  noitemsep,
  font=\normalfont
}
\newcolumntype{R}{>{\raggedleft\arraybackslash}X}

\DeclareMathOperator*{\argmin}{arg\,min}

\newcommand{\ql}{QuantLab}

\newcommand{\citep}[1]{\cite{#1}}
\newcommand{\citet}[1]{\cite{#1}}

\renewcommand\algorithmicindent{0.4em}

\newcommand{\placetextbox}[4]{
  \setbox0=\hbox{#4}
  \AddToShipoutPictureFG*{
    \if#3r
    \put(\LenToUnit{\paperwidth-#1},\LenToUnit{\paperheight-#2}){\vtop{{\null}\makebox[0pt][r]{\begin{tabular}{r}#4\end{tabular}}}}%
    \else
    \put(\LenToUnit{#1},\LenToUnit{\paperheight-#2}){\vtop{{\null}\makebox[0pt][l]{\begin{tabular}{l}#4\end{tabular}}}}%
    \fi
  }%
}%

\newacronym[plural=DVS,firstplural=Dynamic Vision Sensors (DVS)]{dvs}{DVS}{Dynamic Vision Sensor}
\newacronym[plural=SNNs,firstplural=Spiking Neural Networks (SNN)]{snn}{SNN}{Spiking Neural Network}
\newacronym[plural=TCNs,firstplural=Temporal Convolutional Networks (TCN)]{tcn}{TCN}{Temporal Convolutional Network}
\newacronym[plural=DNNs,firstplural=Deep Neural Network (DNN)]{dnn}{DNN}{Deep Neural Network}
\newacronym[plural=CNNs,firstplural=Convolutional Neural Networks]{cnn}{CNN}{Convolutional Neural Network}
\newacronym[plural=BNNs,firstplural=Binarized Neural Networks (BNNs)]{bnn}{BNN}{Binarized Neural Network}
\newacronym[plural=TNNs,firstplural=Ternarized Neural Networks (TNNs)]{tnn}{TNN}{Ternarized Neural Network}
\newacronym[plural=QNNs,firstplural=Quantized Neural Networks (QNNs)]{qnn}{QNN}{Quantized Neural Network}
\newacronym{dnas}{DNAS}{differentiable neural architecture search}
\newacronym{soa}{SoA}{state of the art}
\newacronym{mac}{MAC}{multiply-accumulate}
\newacronym{inq}{INQ}{Incremental Network Quantization}
\newacronym{ste}{STE}{Straight-Through-Estimator}
\newacronym{bn}{BN}{Batch Normalization}
\newacronym{soc}{SoC}{System-on-Chip}
\newacronym{simd}{SIMD}{Single Instruction Multiple Data}
\newacronym{fq}{FQ}{fake-quantized}
\newacronym{ptq}{PTQ}{Post-Training Quantization}
\newacronym{qat}{QAT}{Quantization-Aware Training}
\newacronym{tqt}{TQT}{Trained Quantization Thresholds}
\newacronym{mcu}{MCU}{microcontroller unit}
\newacronym{lsq}{LSQ}{Learned Step Size Quantization}
\newacronym{pact}{PACT}{Parametrized Activation Clipping}
\newacronym{htanh}{HtanH}{Hard Hyperbolic Tangent}
\newacronym{cv}{CV}{cross-validation}
\newacronym[plural=BOPs, firstplural=binary operations (BOPs)]{bop}{BOP}{binary operation}
\newacronym{pfc}{FC}{fabric controller}
\newacronym{fc}{FC}{fully-connected}
\newacronym{mnv1}{MNv1}{MobileNetV1}
\newacronym{mnv2}{MNv2}{MobileNetV2}
\newacronym{iot}{IoT}{internet of things}
\newacronym{dl}{DL}{deep learning}
\newacronym{rl}{RL}{reinforcement learning}
\newacronym[plural=ISAs, firstplural=instruction set architectures (ISAs)]{isa}{ISA}{instruction set architecture}
\newacronym{pulp}{PULP}{parallel ultra-low-power}
\newacronym{ppo}{PPO}{proximal policy optimization}
\newacronym{ddpg}{DDPG}{deep deterministic policy gradient}
\newacronym{icn}{ICN}{integer channel norm}
\newacronym{ilp}{ILP}{integer linear programming}
\newacronym[plural=LTs, firstplural=layer types (LTs)]{lt}{LT}{layer type}
\newacronym{mp}{MP}{mixed-precision}

\def\BibTeX{{\rm B\kern-.05em{\sc i\kern-.025em b}\kern-.08em
    T\kern-.1667em\lower.7ex\hbox{E}\kern-.125emX}}
\begin{document}
\placetextbox{0.5cm}{0.5cm}{l}{\parbox{20cm}{\footnotesize This paper has been accepted at IEEE AICAS
  2023. \copyright 2023 IEEE. Personal use of this material is permitted. Permission from IEEE must be obtained for all other uses, in any current or future media, including reprinting/republishing this material for advertising or promotional purposes, creating new collective works, for resale or redistribution to servers or lists, or reuse of any copyrighted component of
  this work in other works.}}
\title{Free Bits: Latency Optimization of Mixed-Precision Quantized Neural Networks on the Edge
  \thanks{This work is funded in part by the Convolve project evaluated by the EU Horizon Europe research and innovation programme under grant agreement No. 101070374 and has been supported by the Swiss State Secretariat for Education Research and Innovation under contract number 22.00150.}
}

\author{\IEEEauthorblockN{Georg Rutishauser\IEEEauthorrefmark{1}, Francesco Conti\IEEEauthorrefmark{2}, Luca Benini\IEEEauthorrefmark{1}\IEEEauthorrefmark{2}}
\IEEEauthorblockA{\IEEEauthorrefmark{1}\textit{Departement Informationstechnologie und Elektrotechnik, ETH
    Z{\"u}rich, Switzerland}}
\IEEEauthorblockA{\IEEEauthorrefmark{2}\textit{Dipartimento di Ingegneria dell'Energia Elettrica e
    dell'Informazione, Universit\`{a} di Bologna, Bologna, Italy}}
\IEEEauthorblockA{\IEEEauthorrefmark{1}\texttt{\{georgr,lbenini\}@iis.ee.ethz.ch} \IEEEauthorrefmark{2}\texttt{f.conti@unibo.it}}
 }

\maketitle
\begin{abstract}
Mixed-precision quantization, where a deep neural network's layers are quantized to different precisions, offers the opportunity to optimize the trade-offs between model size, latency, and statistical accuracy beyond what can be achieved with homogeneous-bit-width quantization.
To navigate the intractable search space of mixed-precision configurations for a given network, this paper
proposes a hybrid search methodology.
It consists of a hardware-agnostic differentiable search algorithm followed by a hardware-aware heuristic
optimization to find mixed-precision configurations latency-optimized for a specific hardware target. We
evaluate our algorithm on MobileNetV1 and MobileNetV2 and deploy the resulting networks on a family of
multi-core RISC-V microcontroller platforms with different hardware characteristics. We achieve up to \SI{28.6}{\percent}
reduction of end-to-end latency compared to an 8-bit model at a negligible accuracy drop from a full-precision
baseline on the 1000-class ImageNet dataset. We demonstrate speedups relative to an 8-bit baseline, even on
systems with no hardware support for sub-byte arithmetic at negligible accuracy drop. Furthermore, we show the
superiority of our approach with respect to differentiable search targeting reduced binary operation counts as
a proxy for latency. 
\end{abstract}


\section{Introduction}
\label{sec:introduction}
The number of \gls{iot} devices deployed is growing rapidly and is projected to reach 19.1 billion by 2025
\cite{ref:statista_report}. 
To efficiently and accurately process the massive amounts of data collected by \gls{iot} sensor nodes under
the strict latency and power constraints imposed by \gls{iot} applications, the emerging field of Edge AI aims
to deploy \gls{dl} algorithms directly on the edge devices that collect them. \Glspl{mcu} have been a popular
target for edge deployment of \glspl{dnn} due to their ubiquity and low cost, and extensive research has been
conducted into designing efficient \gls{dnn} models and techniques to enable inference on \glspl{mcu}
\cite{ref:mcunetv2,ref:micronets}. As the active power consumption of edge nodes is generally dominated by
components other than the arithmetic units, the most effective way to decrease the full-system inference
energy on a given system is by reducing the inference latency while meeting their tight memory and storage
constraints.

A key technique to reduce both memory footprint and inference latency of \glspl{dnn} is \textit{quantization},
where model parameters and intermediate activations are represented in low-precision formats. 8-bit quantized
models generally exhibit equivalent accuracy to full-precision models. Thanks to \gls{simd} instructions,
these models can be executed on modern \gls{mcu}-based system with lower latency and correspondingly reduced
energy cost~\cite{ref:armv8.1}.
Quantization to even lower bit-widths has also seen widespread interest~\cite{ref:pact_sawb, ref:inq, ref:tnn}
and the hardware community has followed suit, proposing low-precision \gls{dnn} execution engines as well as
\gls{isa} extensions to accelerate networks quantized to sub-byte
precision~\cite{ref:dustin}.

However, homogeneous quantization to sub-byte precisions often
incurs a non-negligible accuracy penalty. To find the best trade-off between execution latency and statistical
accuracy, \textit{mixed-precision quantization} proposes to quantize different parts (usually at the
granularity of individual layers) of the network to different precisions.
In order to efficiently navigate the intractable search space of precision configurations of a given model,
multiple works have applied \gls{dnas} to mixed-precision search. These approaches generally rely on a proxy
for latency, such as \gls{bop} count, to guide the search \cite{ref:bb,ref:edmips,ref:bitprune}. By modeling
quantization to different precisions in a differentiable manner and adding a regularizer term to the loss
to penalize high \gls{bop} counts, these algorithms jointly minimize networks' \gls{bop} count and task
accuracy. The trade-off between operational complexity and accuracy is controlled by the regularization
strength.

\begin{figure*}[t]
    \centering
    \includegraphics[width=\textwidth]{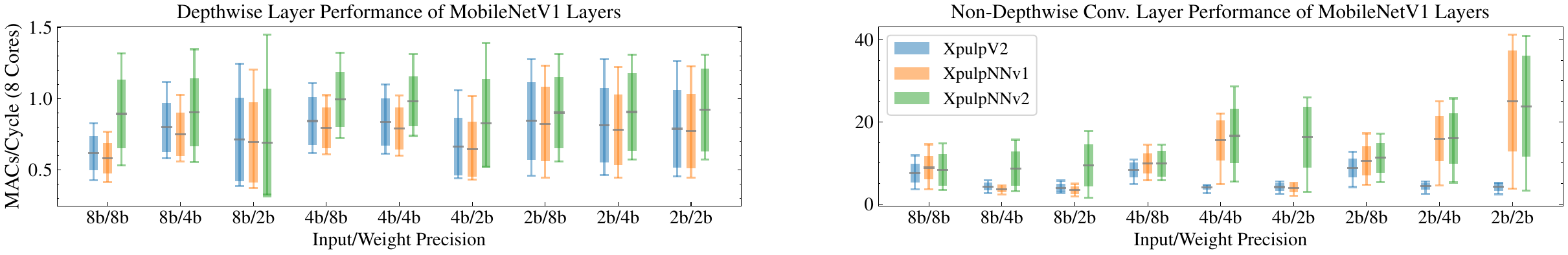}
\caption{Throughput vs. precision of \gls{mnv1} layers on systems implementing XpulpV2, XpulpNNV1 and XpulpNNV2 \gls{isa}
  extensions. 
  Candlesticks' bodies are centered around the mean throughput and extend to one standard deviation above and below it. The ends of the wicks represent the minimum and maximum throughput of all layers.}
\label{fig:freebie_effects}
\end{figure*}

However, as \Cref{fig:freebie_effects} shows, low \gls{bop} counts do not directly translate to reduced
execution latency on real hardware platforms, which motivates our work. Depending on the target platform,
certain layers even exhibit a higher latency when quantized to lower precisions. One reason for this
counterintuitive behavior is in the hardware implementation of low-precision operations. E.g., the XpulpNN
\gls{isa} extension used by \cite{ref:dustin} only supports operands of equal precision - when weight and
activation precisions differ, the lower-precision operands must be unpacked to the larger data
format. 
In systems with hierarchically organized memory, tiling effects also shape the precision-latency landscape:
larger tiles lead to more efficient execution, with the consequence that some layers exhibit
lower latency in sub-8-bit precision on systems without hardware support for sub-8-bit arithmetic (XPulpV2 in
\Cref{fig:freebie_effects}). Past works have targeted the search for mixed-precision networks for deployment
to \gls{mcu}-class platforms. \cite{ref:rusci_mixed} focuses on reducing the memory/storage footprints, but
did not consider inference latency. \cite{ref:channelwise_mp_quant} targets inference energy reduction with a
DNAS-based approach to channel-wise quantization of convolutional layers. However, the channel-wise approach
requires a more complex runtime to enable deployment, and we achieve equivalent results with layer-wise
quantization on a much more challenging dataset than the MLPerf Tiny benchmark tasks used in
\cite{ref:channelwise_mp_quant}.

In this paper, \textit{we propose a mixed-precision latency optimization method consisting of a hardware-agnostic %
differentiable search step
followed by a hardware-aware, profiling-based heuristic which both reduces execution latency and improves
accuracy by increasing the precision in layers where higher precisions achieve lower latency.} We use
Bayesian Bits \cite{ref:bb} for the first step, but our method is not specific to Bayesian Bits and
any mixed-precision search method could be used in its stead.

In evaluations on \gls{mnv1} and \gls{mnv2}, deployed to a cycle-accurate RISC-V multi-core simulator, our
approach results in an accuracy-latency trade-off curve that dominates those produced by differentiable search
alone. To the best of our knowledge, we demonstrate for the first time end-to-end deployment of mixed-precision
networks to an \gls{mcu}-class platform that exhibit not only a reduced memory footprint but also reduced
execution latency by up to \SI{28.6}{\percent} at full-precision equivalent classification accuracy.
Our key contributions are the following:
\begin{itemize}
    \item We present a lightweight method to find latency-optimized mixed-precision quantization configurations for \glspl{dnn}, consisting of a hardware-agnostic differentiable model search and hardware-aware heuristics, allowing efficient generation of optimized configurations for different platforms.
    \item We compose an end-to-end flow consisting of precision search, training, generation of integerized models and deploy the found configurations on a cycle-accurate simulator for high-performance RISC-V \gls{mcu} systems.
    \item We analyze the resulting accuracy-latency trade-offs, demonstrating a reduction of end-to-end latency by up to \SI{28.6}{\percent} vs. 8-bit quantization at full-precision equivalent classification accuracy and finds Pareto-dominant configurations with respect to homogeneous 4-bit quantization.
\end{itemize}
\section{Free Bits}
\label{sec:algo}
\textit{Free Bits} is a multi-step method to find mixed-precision configurations of \glspl{dnn} optimized for
low latency on a given target hardware platform. In the first step, we employ two variants of the Bayesian
Bits algorithm 
to find baseline reduced-precision configurations of the
targeted network architecture.

In the second step, we use layer-wise profiling data collected on the target platform to update the initial
configuration by increasing the precision of layers which exhibit lower latency in higher precisions.
As these increases in precision are expected to improve both latency
and statistical accuracy, we name this step the \textit{free bits} heuristic.


From the configurations found in the second step, the one which best meets a given latency target is then
selected and fine-tuned using a modified version of the \gls{tqt} algorithm \cite{ref:tqt} and automatically
converted to an integer-only model, which can be fed to a deployment backend 
for the target platform.


\subsection{Differentiable Mixed-Precision Search}
\label{subsec:diff_search}
To find the initial mixed-precision configurations,
we apply two variants of the Bayesian Bits algorithm. Bayesian Bits decomposes the quantization of each layer
into the contributions from each of the allowed bit-widths and aims to reduce a network's total \gls{bop}
count with a regularizer that penalizes each precision's contribution to the expected \gls{bop} count
individually. Because the execution latency of a layer does not necessarily increase monotonously with
precision and depends jointly on input and weight precisions, Bayesian Bits cannot target latency reduction
directly.
In addition to the original Bayesian Bits algorithm, we also employ a modified version 
enforcing equal input and weight precisions. This modification accounts for the fact that on our target
platforms, the theoretical throughput for a layer with non-equal activation and weight precisions is bounded
by the higher of the two precisions.

\subsection{Free Bits Heuristic}
\label{subsec:profiling_heuristic}
\begin{algorithm}
\caption{Free Bits Heuristic}
    \label{algo:freebie_heur}
    \hspace*{\algorithmicindent}\textbf{Input:}
    \begin{AlgoDx}
        \item[$LD$:] Latency dictionary mapping \acrshort{lt} $t$ and precisions $(b_{in}, b_{wt})$ to a measured latency
        \item[$C_{net}$:] Dictionary of \acrshortpl{lt} and precisions describing a \acrshort{mp} network, of the form $\left\{i:\left(t^i, \left( b_{in}^i, b_{wt}^i\right)\right)\right\}_{i=1}^N$
        \item[$P_{all}$:] Set of allowed combinations $\left( b_{in}, b_{wt}\right)$ of input and weight precisions
    \end{AlgoDx}
    \hspace*{\algorithmicindent}\textbf{Output:}
    \begin{AlgoDx}
        \item[$C'_{net}$:] Latency-optimized \acrshort{mp} configuration of input network
    \end{AlgoDx}
    \begin{algorithmic}
    \Function{higher}{$(b_{in, 1}, b_{wt,1}), (b_{in,2}, b_{wt,2})$}
    \State \textbf{return} $(b_{in,1} \geq b_{in,2}) \wedge (b_{wt,1} \geq b_{wt,2})$
    \EndFunction
    \State $C'\gets C$
    \ForAll{$i, c^i=\left(t^i, \left( b_{in}^i, b_{wt}^i\right)\right)\in C_{net}$}
    \State $lat^i_0\gets LD\left[ c^i\right]$
    \Comment{Initial latency}
    \State $cdts\gets$\parbox[t]{0.5\linewidth}{$\{ \left(b_{in}, b_{wt}\right)\vert $ $\left(b_{in},
        b_{wt}\right)\in P_{all},$ $LD\left[\left(t^i, \left(b_{in},b_{wt}\right)\right)\right]\leq
      lat^i_0,$

    $\Call{higher}{(b_{in}, b_{wt}), (b_{in}^i,
      b_{wt}^i)}\}$}\Comment{\parbox[t]{0.27\linewidth}{\linespread{0.93}\selectfont Select lower-lat. configs with higher precisions}}
  \State $best\gets$ \parbox[t]{0.5\linewidth}{$ \argmin\limits_{(b_{in},b_{wt})\in cdts}LD[(t^i,
    (b_{in},b_{wt}))]$}\Comment{\parbox[t]{0.25\linewidth}{\linespread{0.93}\selectfont Select lowest-lat. candidate}}
    \State $C'_{net}[i]\gets (t^i, best)$
    \Comment{Update net configuration}
    \EndFor
    \end{algorithmic}
  \end{algorithm}
As \Cref{fig:freebie_effects} shows, there are many cases where a given layer does not profit from reduced
precision, but in fact exhibits higher latency when executed in a lower precision. The \textit{free bits}
heuristic exploits this observation, relying on two core ideas: First, we assume our target platform executes
networks layer-by-layer, which implies $L_{net} \approx \sum_{i=1}^NL_i$ for the total execution latency
$L_{net}$ of an $N$-layer network where the $i$-th layer is executed with latency $L_i$.
Second, increasing a layer's input activation or weight precision never decreases the network's statistical
accuracy.

Following the first assumption, we characterize each unique linear operator in the target network as a
\textit{\gls{lt}}, the tuple of all quantities that parametrize the invocation of a
computational kernel, such as input dimensions, number of channels, or kernel size.
For each \gls{lt} occurring in the network, we profile the execution latency on the target platform for all
supported precision configurations.
We then update every layer in the network found by Bayesian Bits to the configuration
of higher or equal precision that exhibits the lowest latency.
By the two assumptions above, the resulting network's execution latency and statistical accuracy will be
upper-bounded and lower-bounded, respectively, by those of the configuration found by Bayesian Bits. As it is
expected to produce strictly superior configurations in terms of latency and statistical accuracy, we call
this procedure the \textit{free bits} heuristic. We show a pseudocode description of the procedure in
\Cref{algo:freebie_heur}.

\subsection{Quantization-Aware Fine-Tuning and Deployment}
Having arrived at a latency-optimized mixed-precision configuration, we perform \gls{qat} with the \ql
framework \footnote{\url{https://github.com/pulp-platform/quantlab/tree/georgr/bayesian_bits_gh}} to fine-tune the
network's parameters using a generalized version of \gls{tqt}~\cite{ref:tqt}, differing from the original
algorithm in that we do not force clipping bounds to be exact powers of two.

\section{Results}
\label{sec:results}
\subsection{Experimental Setup}
\label{subsec:exp_setup}
We performed experiments on the well-known and widely used MobileNetV1 and V2 architectures, applying the procedure proposed in \Cref{sec:algo} to \gls{mnv1} \cite{ref:mnv1} and \gls{mnv2} \cite{ref:mobilenetv2}. We used width multipliers of 0.75 for \gls{mnv1} and 1.0 for \gls{mnv2}. The input resolution was $224\times 224$ for both networks. We trained our networks on the ILSVRC2012 \cite{ref:imagenet} 1000-class dataset and report top-1 classification accuracies on the validation set.
\paragraph{Differentiable Mixed-Precision Search and \gls{qat} Fine-Tuning}
We applied the two variants of Bayesian Bits described in \Cref{subsec:diff_search} to the \gls{mnv1} and
\gls{mnv2} network topologies. 
The configurations produced by our algorithm (as well as those produced by Bayesian Bits in the case of
\gls{mnv1}) were fine-tuned with \gls{tqt}.
In accordance with the capabilities of our hardware targets (see below), the precisions Bayesian Bits can select from are 2, 4, and 8 bits for both weights and activations. 

\paragraph{Profiling, Deployment and Hardware Targets}
\ql's automated integerization flow generates precision-annotated, integer-only ONNX models, which are
consumed by the DORY \cite{ref:dory} deployment backend. DORY generates C code leveraging a
mixed-precision kernel library, which we run on GVSOC, a cycle-accurate, open-source simulator for multi-core
RISC-V systems. The platforms we target are open-source RISC-V \glspl{mcu} of the \gls{pulp} family and are
divided into two main domains. The \gls{soc} domain contains a RISC-V core serving as the fabric controller,
\SI{512}{\kibi\byte} of L2 memory and a full set of peripherals. The cluster domain hosts 8 high-performance
RISC-V cores operating on \SI{64}{\kibi\byte} of high-bandwidth L1 scratchpad memory.
This hierarchical memory structure necessitates \textit{tiled} execution of a network's layers with each
tile's inputs, outputs, and weights fitting into the L1 scratchpad. Tiling is automatically performed by
DORY. 

All cores in the target system implement the base RV32IMF \gls{isa} and the custom XpulpV2
extensions. We measure latency on three systems, with varying degrees of support for sub-byte arithmetic in
the cluster cores. The \textit{XpulpV2} system's cluster implements only XpulpV2, which supports only 8-bit
\gls{simd} arithmetic. The \textit{XpulpNNv1} system implements the XpulpNN extension also used in \cite{ref:dustin},
which provides support for packed-SIMD sub-byte arithmetic on 2- and 4-bit data. 
 Because XpulpNN's arithmetic instructions require operands to have equal bit-width, mismatching activation
 and weight precisions require lower-precision data to be unpacked in software to the higher precision.
 Finally, the \textit{XpulpNNv2} system eliminates this overhead by performing the unpacking transparently in
 Hardware. To generate the profiling data (shown in \Cref{fig:freebie_effects} for \gls{mnv1}) used by the free bits
 heuristic, we again use DORY to generate and export dummy networks for all layer types in all precision
 configurations. 
\subsection{Latency-Accuracy Trade-Offs for XpulpNNv1}
\begin{figure}
    \centering
    \includegraphics[width=\linewidth]{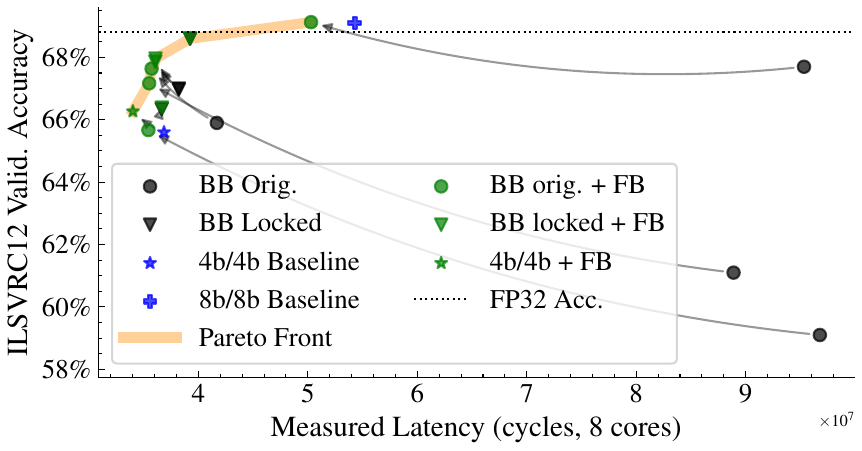}
    \caption{Latency-Accuracy tradeoff of Bayesian Bits-trained MobileNetV1 configurations before and after
      applying the free bits heuristic. Grey arrows indicate the effect of the heuristic. 
      \textbf{BB Orig./Locked}: Configurations found by the original Bayesian Bits algorithm and the modified version enforcing symmetric activation/weight precisions, respectively. \textbf{FB}: Free bits}
    \label{fig:bb_and_heur}
\end{figure}
\paragraph{MobileNetV1}
\Cref{fig:bb_and_heur} shows the latency-accuracy trade-off for \gls{mnv1} deployed to a \gls{pulp} system
with the XpulpNNv1 \gls{isa} extensions, with the effect of the free bits heuristic indicated. We observe that
the original Bayesian Bits algorithm generally does not produce low-latency configurations due to the reasons
discussed in \Cref{sec:introduction}.
With two exceptions, applying the free bits heuristic improves the latency of all configurations substantially
while increasing classification accuracy. For the $4\,$b/$4\,$b baseline, the heuristic
increases the precision of 12 layers, improving latency and accuracy by $7\%$ and $0.7$ percentage points,
respectively. As symmetric activation and weight precisions are theoretically optimal for XpulpNNv1's hardware
implementation of sub-byte arithmetic, this is a non-trivial result. The free bits heuristic lifts the
previously uncompetitive configurations found by the original Bayesian Bits algorithm to the Pareto front,
yielding accuracy and latency gains of $1.4-6.6$ percentage points and $12.3\%-61.6\%$, respectively. The
most accurate configuration matches the 8b/8b baseline in statistical accuracy at $69.1\%$ and reduces
execution latency by $7.6\%$, and the configuration at the Pareto front's knee point improves execution
latency by $27.9\%$ at a classification accuracy within $0.2$ percentage points of the full-precision 
baseline of $68.8\%$.
\paragraph{MobileNetV2}
\begin{figure}
    \centering
    \includegraphics[width=\linewidth]{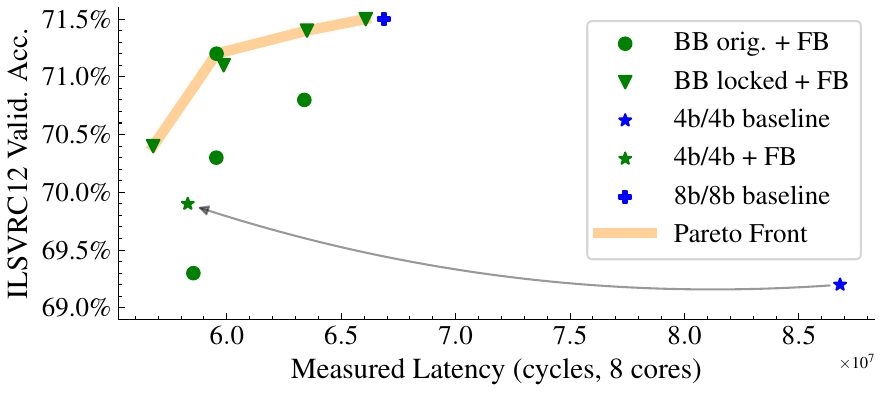}
    \caption{Latency-Accuracy tradeoff of MobileNetV2 configurations optimized for XpulpNNv1. The grey arrow
      indicates the effect of the heuristic on the $4$b/$4$b baseline.}
    \label{fig:mnv2_pareto}
\end{figure}
\Cref{fig:mnv2_pareto} shows the latency-accuracy trade-off of \gls{mnv2} configurations produced by Bayesian
Bits modified with the free bits heuristic running on the XpulpNNv1 system. The baseline 4b/4b configuration
contains many asymmetric-precision convolutional layers due to adder node outputs being quantized to 8 bits.
This leads to a latency higher than that of the 8b/8b baseline, which the free bits heuristic reduces by
$46\%$ while improving classification accuracy by $0.6$ percentage points. Nevertheless, the
resulting configuration is not Pareto-optimal with respect to those produced by our algorithm. In particular,
the locked-precision version of Bayesian Bits, when combined with the free bits heuristic, produces
configurations that dominate both baselines. The configuration at the
Pareto front's knee point reduces execution latency by $10.9\%$ at an accuracy penalty of only $0.3$
percentage points from the 8b/8b baseline.
\subsection{Free Bits Across Different Target Platforms}
\begin{table}[]
    \centering
    \begin{tabular}{l|lrr|rr}
         Acc. Margin &\gls{isa} & \multicolumn{2}{c}{MobileNetV1} & \multicolumn{2}{|c}{MobileNetV2}  \\\hline\hline
         \multicolumn{1}{c}{}& & Lat. vs. 8b & Acc. & Lat. vs. 8b & Acc. \\\hline
         \multicolumn{1}{l|}{8b Baseline} &\textit{all} & $+0\%$ & $69.1\%$ & $+0\%$ & $71.5\%$\\\hline
       \multirow{3}{*}{$0.5$ pp.}  & XPv2 & $-5.5\%$& $69.3\%$ & $-3.4\%$ & $71.0\%$ \\
       & XPNNv1 & $-27.9\%$ & $68.6\%$ & $-10.9\%$ & $71.2\%$ \\
       & XPNNv2 & $-28.6\%$ & $68.6\%$ & $-15.3\%$ & $71.0\%$ \\\hline
       \multirow{3}{*}{$1.5$ pp.} & XPv2 & $-5.5\%$ & $69.3\%$ & $-6.3\%$ & $70.7\%$\\
       & XPNNv1 & $-34.4\%$ & $67.6\%$ & $-15.1\%$ & $70.4\%$ \\
       & XPNNv2 & $-35.1\%$ & $67.6\%$ & $-15.3\%$ & $71.0\%$\\\hline
       \multirow{3}{*}{4b + FB} & XPv2 & $-3.5\%$ & $67.7\%$ & $-7.7\%$ & $70.9\%$ \\
       & XPNNv1 & $-37.1\%$ & $66.3\%$ & $-12.8\%$ & $69.9\%$\\
       & XPNNv2 & $-39.8\%$ & $66.6\%$ & $-25.7\% $ & $69.6\%$ \\\hline
       \multirow{3}{*}{4b Baseline} & XPv2 & $+49.9\%$ & $65.6\%$ & $+37.0\%$ & $69.3\%$ \\
       & XPNNv1 & $-32.3\%$ & $65.6\%$ & $+48.9\%$ & $69.3\%$ \\
       & XPNNv2 & $-38.3\%$ & $65.6\%$ & $-23.4\%$ & $69.3\%$ \\
    \end{tabular}
    \caption{Configurations within margins of 0.5 and 1.5 percentage points (\textbf{pp.}) of 8b/8b
      classification accuracy for PULP systems implementing different ISA extensions: XpulpV2 (\textbf{XPv2}),
      XpulpNNv1 (\textbf{XPNNv1}) and XpulpNNv2 (\textbf{XPNNv2}). 
      \textbf{4b+FB}: target-specific free bits heuristic applied to homogeneously quantized 4b/4b network.}
    \label{tbl:best_nets}
\end{table}
To evaluate the portability of our algorithm, we optimized  \gls{mnv1} and \gls{mnv2} configurations found
with Bayesian Bits for the three different \gls{pulp} systems 
described in \Cref{subsec:exp_setup}. \Cref{tbl:best_nets} shows the lowest-latency configurations
within $0.5$ and $1.5$ percentage points of classification accuracy of the 8b/8b baseline.
Notably, our approach achieves latency reductions even on the XpulpV2 system without hardware support for sub-byte arithmetic, which can be attributed to a lower data movement overhead thanks to larger tile sizes. 

\section{Conclusion}
In this paper, we have presented \textit{Free Bits}, an efficient method to find latency-optimized
mixed-precision network configurations for inference on edge devices. Taking advantage of the fact that,
depending on the target platform, increasing input or weight precision may lead to lower execution latency,
the method optimizes mixed-precision configurations found by the hardware-agnostic Bayesian Bits
differentiable search algorithm. Deploying the \gls{mnv1} and \gls{mnv2} configurations found with our
algorithm on a family of high-performance \gls{mcu}-class RISC-V platforms, we find that, i) with hardware
support for sub-byte arithmetic, \gls{mnv1} end-to-end latency can be reduced by up to \SI{30}{\percent} while
retaining full-precision equivalent accuracy, ii) even without such hardware support, mixed-precision
quantization enables a latency reduction of up to $7.7\%$, and iii) the found configurations offer a superior
accuracy-latency trade-off with respect to homogeneous 4-bit and 8-bit quantization.

\bibliography{library,armv8}
\end{document}